\documentclass[10pt,twocolumn,letterpaper]{article}

\usepackage[pagenumbers]{cvpr}

\usepackage{graphicx}
\usepackage{amsmath}
\usepackage{amssymb}
\usepackage{booktabs}
\usepackage{array}  
\usepackage{tabularx}
\usepackage[toc,page]{appendix}
\usepackage[dvipsnames]{xcolor}
\usepackage[pagebackref,breaklinks,colorlinks]{hyperref}
\DeclareMathOperator*{\argmax}{argmax}
\def\ie{\emph{i.e}\onedot}

\usepackage[capitalize]{cleveref}
\crefname{section}{Sec.}{Secs.}
\Crefname{section}{Section}{Sections}
\Crefname{table}{Table}{Tables}
\crefname{table}{Tab.}{Tabs.}

\usepackage{capt-of,etoolbox}	
\makeatletter
\apptocmd\@maketitle{{\myfigure{}\par}}{}{}
\makeatother

\begin{document}
\newcommand\myfigure{%
\centering
\vspace*{-0.15in}
}

\title{Exploring Pixel-level Self-supervision for Weakly Supervised
\\Semantic Segmentation}

\author{
\textbf{Sung-Hoon Yoon\thanks{The first two authors contributed equally.}}
\quad
\textbf{Hyeokjun Kweon$^{*}$}
\quad 
\textbf{Jaeseok Jeong\thanks{The other authors contributed equally.}}
\\
\textbf{Hyeonseong Kim$^{\dagger}$}
\quad
\textbf{Shinjeong Kim$^{\dagger}$}
\quad
\textbf{Kuk-Jin Yoon}
% \vspace{2pt}
\and
Visual Intelligence Lab., KAIST, Korea\\
{\tt\small \{yoon307, 0327june, jason.jeong, brian617, aakseen, kjyoon\}@kaist.ac.kr}
}
\maketitle

\begin{abstract}
   Existing studies in weakly supervised semantic segmentation (WSSS) have utilized class activation maps (CAMs) to localize the class objects.
   However, since a classification loss is insufficient for providing precise object regions, CAMs tend to be biased towards discriminative patterns (i.e., sparseness) and do not provide precise object boundary information (i.e., impreciseness).
   To resolve these limitations, we propose a novel framework (composed of MainNet and SupportNet.) that derives pixel-level self-supervision from given image-level supervision.
   In our framework, with the help of the proposed Regional Contrastive Module (RCM) and Multi-scale Attentive Module (MAM), MainNet is trained by self-supervision from the SupportNet.
   The RCM extracts two forms of self-supervision from SupportNet: (1) class region masks generated from the CAMs and (2) class-wise prototypes obtained from the features according to the class region masks.
   Then, every pixel-wise feature of the MainNet is trained by the prototype in a contrastive manner, sharpening the resulting CAMs.
   The MAM utilizes CAMs inferred at multiple scales from the SupportNet as self-supervision to guide the MainNet. Based on the dissimilarity between the multi-scale CAMs from MainNet and SupportNet, CAMs from the MainNet are trained to expand to the less-discriminative regions.
   The proposed method shows state-of-the-art WSSS performance both on the train and validation sets on the PASCAL VOC 2012 dataset. 
   For reproducibility, code will be available publicly soon.
\end{abstract}

\vspace{-10pt}
\section{Introduction} \label{sec:intro}
Owing to the development of deep learning, the performance of semantic segmentation~\cite{zhu2019improving,yuan2020object,sun2019high,cheng2020panoptic,takikawa2019gated,li2020improving} algorithms has been improved rapidly over the recent few years.
However, most recent deep-learning based algorithms have become reliant on large labeled datasets to achieve superior performance. 
One of the key challenges for semantic segmentation research lies in the fact that it is difficult to obtain dense pixel-level labeled datasets to train the deep-learning networks.
This deficiency in labeled datasets is much more apparent in semantic segmentation than in other vision tasks, such as image classification or object detection.

\begin{figure}
    \centering
    \includegraphics[width=1\linewidth]{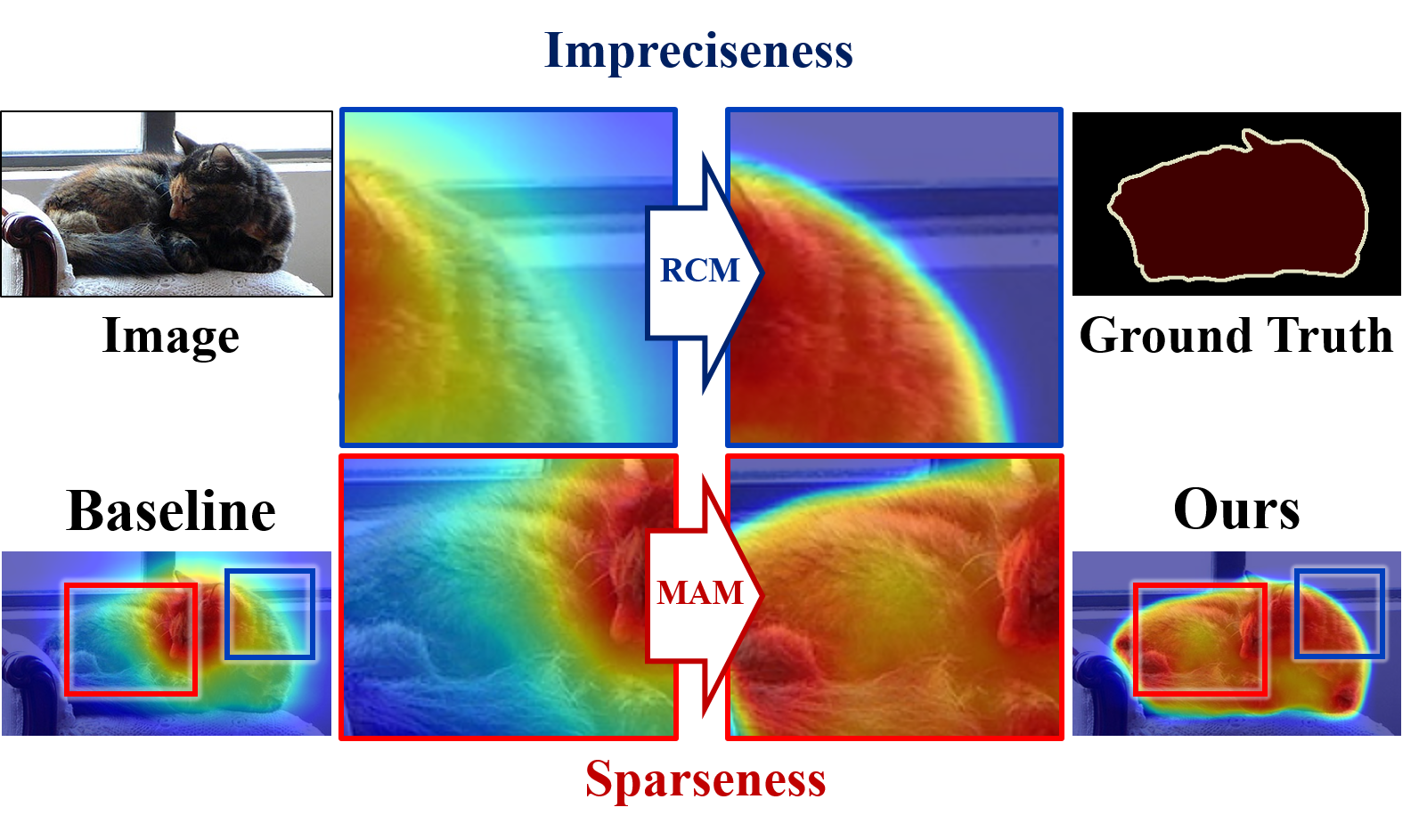}
    \vspace{-23pt}
    \caption{Limitations of CAMs that our framework targets: the Regional Constrastive Module (RCM) addresses the impreciseness and the Multi-scale Attention Module (MAM) aims to relieve the sparseness.}
    \vspace{-15pt}
    \label{fig:intro_concept}
\end{figure}

To address this problem, weakly supervised semantic segmentation (WSSS) studies have utilized labels containing limited information, such as image-level labels~\cite{ahn2018learning,wang2020self,zhang2020reliability,fan2020learning,ahn2019weakly,chang2020weakly,lee2021anti,kweon2021unlocking,zhang2021complementary,li2021pseudo}, scribbles~\cite{lin2016scribblesup,vernaza2017learning}, and bounding boxes~\cite{dai2015boxsup,khoreva2017simple,papandreou2015weakly,lee2021bbam}, for producing dense pixel-level labels. 
In particular, utilizing only image-level classification labels is the most actively studied approach in WSSS since image-level labels are much more accessible than other labels.
As such, in this paper, we propose a method to utilize image-level classification labels only for WSSS.

The existing studies in WSSS aim to generate pixel-level labels which can serve as pseudo ground truths (GTs) to train a semantic segmentation model.
When using image-level labels, objects are localized generally through class activation maps (CAMs)~\cite{zhou2016learning}. 
Though using CAMs is the most widely used localization technique in image-level label-based WSSS, it also has several limitations.
Semantic segmentation and image-level classification are similar in their semantic nature of the task; however, the supervision required is not interchangeable.
Compared to semantic segmentation, which can be regarded as a pixel-level classification task, the image-level classification task does not require pixel-level supervision.
Therefore, the image classifier learns to predict the correct class for the whole image rather than for each pixel, by focusing on the most discriminative patterns shared among images of the same class. As shown in Fig.~\ref{fig:intro_concept}, the resulting CAMs naturally tend to highlight the discriminative regions of the object (sparseness) and do not fit along the object boundary (impreciseness).

To overcome the aforementioned limitations of CAMs, we propose a framework, which derives pixel-level self-supervision from image-level supervision, to acquire precise CAMs suitable for semantic segmentation.
In our framework, a guided network (MainNet) is optimized through gradient descent while the guidance network (SupportNet) is updated through Exponential Moving Average (EMA) as in~\cite{He_2020_CVPR}.
To produce and impose self-supervision for training MainNet, we devise the Regional Contrastive Module (RCM) and the Multi-scale Attentive Module (MAM), where the RCM mainly resolves impreciseness while the MAM relieves sparseness.

In the RCM, class region masks are generated by thresholding the CAMs from the SupportNet.
These masks are used as pixel-level self-supervision that notifies which pixels belong to which class.
Though we can use the masks as direct supervision for MainNet as in~\cite{zhang2020reliability}, we impose an indirect training scheme based on contrastive learning to prevent MainNet from learning directly from erroneous supervision.
We make the RCM build up representative features, by fusing the features according to the class region mask.
To increase the representative power while relieving the erroneous nature of self-supervision, we aggregate the representative features for each class from all images along the batch dimension and define the resulting vectors as class-wise prototypes.
Then, according to the class region masks, each pixel-level feature of the MainNet is trained to be close to the corresponding class prototype, while being far from the prototypes of the other classes.
Imposing only image-level supervision leads to the imprecise result due to the lack of regional information.
On the other hand, with the pixel-level supervision based on the class region mask and the class-wise prototype, the proposed RCM effectively addresses the impreciseness of the CAMs by guiding which pixel should be classified into which class in a contrastive manner.

Additionally, we tackle the problem of sparse activation through the MAM, which exploits the high localization capability of multi-scale inference~\cite{chen2018encoder,zhao2017pyramid}.
Motivated by the fact that network focuses on different regions depending on the resolution of the input image~\cite{tao2020hierarchical}, we define an attention matrix based on the dissimilarity between the multi-scale CAMs of the MainNet and the SupportNet. 
The corresponding attention scores are used as weighting parameters for generating multi-scale inferred CAMs, which serve as self-supervision to train the MainNet. 
Here, \textit{multi-scale inferred CAMs} (\ie{}, \textit{msinf}-CAMs) are aggregation of CAMs obtained at multiple image scales.

In our framework, the proposed RCM and MAM complement each other.
While the RCM is designed to focus on reducing the impreciseness in CAMs, the MAM reduces sparseness, helping the network see even the less-discriminative regions.
As shown in Fig.~\ref{fig:intro_concept}, this combination of modules enables our framework to generate high-quality CAMs, relieving both impreciseness and sparseness. 
We summarize the contributions of our work as follows: \vspace{-4pt}
\begin{list}{$\bullet$}{\leftmargin=1em \itemindent=0em}
\setlength\itemsep{-3pt}
\item We propose the Regional Contrastive Module (RCM) that relieves impreciseness of CAMs through pixel-level self-supervision, with the generated class region mask and class-wise prototype, in a contrastive manner.
\item We propose the Multi-scale Attentive Module (MAM) that leverages high localization capability of \textit{msinf}-CAMs with the dissimilarity of CAMs at multi-scale, to expand CAMs to less-discriminative regions.
\item We achieve state-of-the-art WSSS performance on both the
PASCAL VOC 2012 \textit{validation} and \textit{test} sets using only the image-level classification labels.
\end{list}
\vspace{-4pt}
\begin{figure*}[t] 
    \centering
    \includegraphics[width=1.0\linewidth]{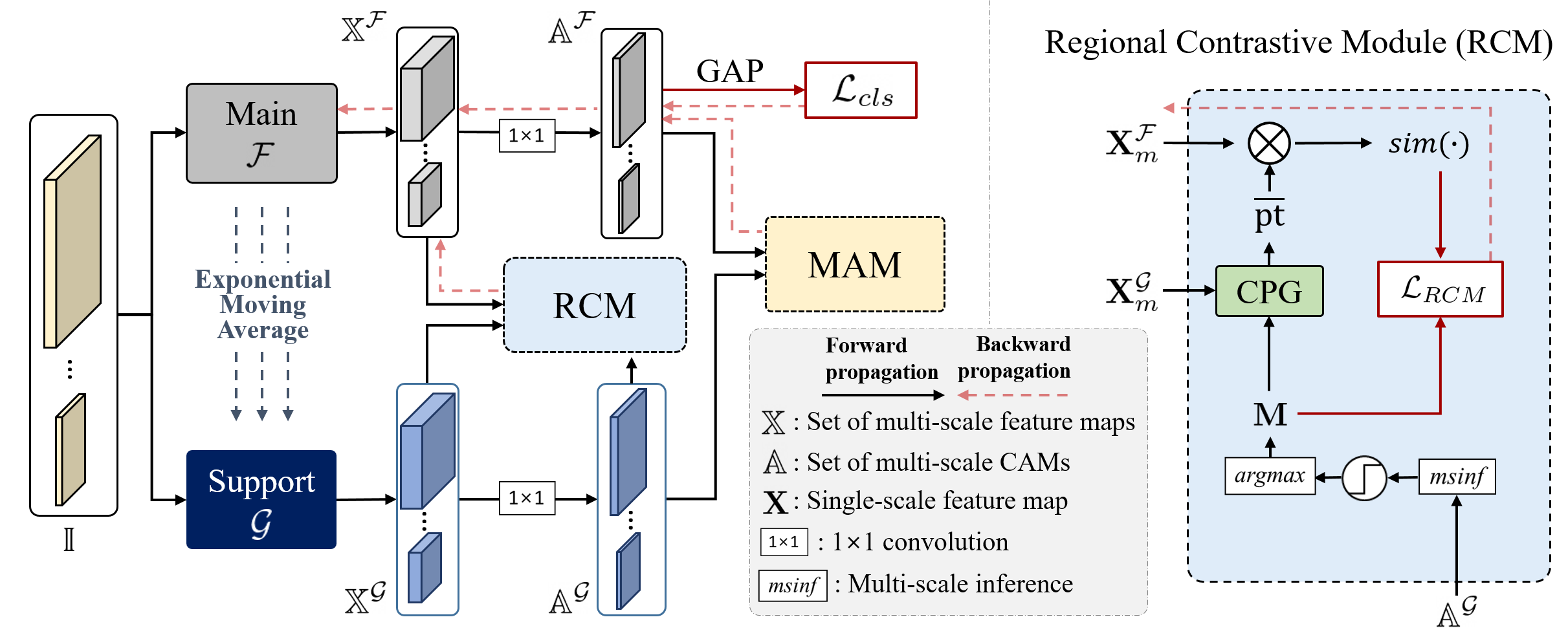}
    \vspace{-10pt}
    \caption{\textbf{\textit{Left.}} Visualization of our proposed framework. The multi-scale inputs are passed to the MainNet $\mathcal{F}$ and SupportNet $\mathcal{G}$. The multi-scale feature maps $\mathbb{X}$ and CAMs $\mathbb{A}$ from the MainNet $\mathcal{F}$ and SupportNet $\mathcal{G}$ are passed into the RCM and MAM, and in return the modules train the MainNet. The SupportNet is updated through EMA following the MainNet.
    \textbf{\textit{Right.}} Visualization of the proposed Regional Contrastive Module (RCM). Class-wise Prototype Generation (CPG) helps provide prototypes for self-supervision.} 
    \label{fig:method_framework}
    \vspace{-10pt}
\end{figure*}

\section{Related Works}

\noindent\textbf{Weakly supervised semantic segmentation} 
To the best of our knowledge, most WSSS studies utilize a classification network to extract CAMs and generate pixel-level pseudo labels using these CAMs.
While using CAMs is an effective way to localize the object in an image, CAMs tend to be biased towards discriminative patterns and have limitations in that they do not provide precise boundary information, which is crucial for semantic segmentation.
To address these limitations, several methods~\cite{kolesnikov2016seed, huang2018weakly} have been proposed to expand and refine the CAMs by constraining the CAMs to fit into object boundaries. Affinity-based methods ~\cite{ahn2018learning,fan2020cian} train a network to learn the pixel-level affinity and apply random walk as post-processing.
By guiding the network to keep searching for objects even after the most discriminative region is erased from the image, Adversarial Erasing (AE) methods~\cite{wei2017object, zhang2018adversarial, hou2018self, li2018tell, kweon2021unlocking} enlarge the CAMs to less-discriminative regions. 
Various techniques such as multi-dilated convolution~\cite{wei2018revisiting}, stochastic feature selection~\cite{lee2019ficklenet}, integration at multiple phases~\cite{jiang2019integral}, and context decoupling augmentation~\cite{su2021context}  are designed to make the network generate better CAMs.
As an additional guidance for network to pay attention to the entire region of objects, some existing works attempt to devise auxiliary tasks such as sub-category classification~\cite{chang2020weakly}, self-equivariant regularization with scale variance minimization~\cite{wang2020self}, class-wise co-attention extraction~\cite{sun2020mining, li2020group}, anti-adversarial attack~\cite{lee2021anti}, and complementary patch loss~\cite{zhang2021complementary}.
Many WSSS methods~\cite{jiang2019integral,fan2020employing,sun2020mining,fan2020learning,li2020group,yao2021non,xu2021leveraging,lee2021railroad} have been proposed to employ the pre-trained saliency detection module, which distinguishes dominant foreground object from its background, as a complementary source of information for enhancing CAMs and generating precise pseudo-pixel labels.
Even though the saliency detection module can help generate high-quality pseudo-pixel labels, to conform the goal of WSSS with the image-level labels only, we neither use the saliency module nor the external dataset.

\noindent\textbf{Self-supervised Learning}
In recent years, without being reliant on massive labeled datasets, self-supervised learning has shown promising results for resolving computer vision tasks such as image denoising~\cite{quan2020self2self,xu2020noisy,laine2019high}, image-level classification~\cite{He_2020_CVPR,chen2020improved,chen2020simple,chen2020big}, and semantic segmentation tasks~\cite{shimoda2019self,wang2020self,zhang2020reliability,zhan2018mix,zhang2020self,van2021unsupervised}.
Self-supervised learning is widely used for feature embedding due to its effective representation capability.
A group of researches~\cite{He_2020_CVPR,chen2020improved,chen2020simple,chen2020big} has been conducted to address image-level classification with self-supervised learning in a contrastive manner.
They aim to obtain feature embedding of the image by minimizing the distance between the features of semantically similar images while that of the different images is maximized.
Recently, some works employed contrastive learning for unsupervised semantic segmentation by grouping the pixels with the help of hierarchical segmentation algorithm~\cite{zhang2020self} or off-the-shelf saliency module~\cite{van2021unsupervised}.
Inspired by these approaches, to resolve the limitations of CAMs in the perspective of WSSS, we propose a method that exploits the CAMs themselves as self-supervision for regional information.
The proposed framework will be discussed in the following section.

\section{Proposed Method}
To address the weaknesses of CAMs for WSSS, we propose a novel framework that derives pixel-level self-supervision from image-level supervision.
We further detail our framework in this section.
Note that our method uses neither saliency detection module nor GT supervision except image-level supervision.

\subsection{Overall framework}
As shown in the left of Fig.~\ref{fig:method_framework}, the proposed framework consists of two structurally identical networks: MainNet($\mathcal{F}$) and SupportNet($\mathcal{G}$).
Self-supervision generated by SupportNet, with the help of the proposed modules (RCM and MAM), is used for training the MainNet.
The gradients obtained through back-propagation only optimize MainNet, while the SupportNet is updated through EMA from the MainNet as~\cite{He_2020_CVPR}.
We denote the weights of MainNet and SupportNet as $\theta_{\mathcal{F}}$ and $\theta_{\mathcal{G}}$, respectively. $\theta_{\mathcal{G}}$ is updated as: $\theta_{\mathcal{G}} \leftarrow \alpha\theta_{\mathcal{G}}+ (1-\alpha)\theta_{\mathcal{F}}$.
With this strategy, we make SupportNet provide stable self-supervision for training the MainNet. Here, $\alpha$ is momentum weight.

Throughout this paper, we will denote the data pair as $(\mathbf{I}, \mathbf{t})$, where $\mathbf{I}$ represents the input image and $\mathbf{t}=\{c_1, c_2, \cdots, c_{N}\}$ represents the classification label in the form of a binary multi-hot vector. 
Note that there are $N$ classes in a set of the foreground classes $C_{fg}=\{1,\cdots,N\}$, and we assign the number $0$ for the background class.
As our method utilizes multi-scale images, we retrieve three different scales of image $\mathbf{I}$ with ratios 0.5, 1.0, and 2.0, represented by the symbols $\mathbf{I}_{s}$, $\mathbf{I}_{m}$, and $\mathbf{I}_{l}$, respectively.
$\mathbf{I}_{s}$ and $\mathbf{I}_{l}$ are obtained through bilinear interpolation of image $\mathbf{I}_{m}$, where $\mathbf{I}_{m}$ is the original image $\mathbf{I}$.
The set of these multi-scale images is represented by $\mathbb{I} = \{\mathbf{I}_s, \mathbf{I}_{m}, \mathbf{I}_{l}\}$ throughout the paper.

When an image $\mathbf{I}$ is given to the networks, a feature map $\mathbf{X} \in \mathbb{R}^{D\times H \times W}$ can be acquired right before the classification head.
By applying a 1$\times$1 convolution to the feature map $\mathbf{X}$, we can formulate CAMs $\mathbf{A}$.
The class prediction is obtained by applying global average pooling (GAP) on the CAMs with sigmoid activation function.

By propagating the multi-scale images $\mathbb{I}$ to both MainNet $\mathcal{F}$ and SupportNet $\mathcal{G}$, a set of multi-scale CAMs from each network denoted by $\mathbb{A}^{\mathcal{F}} = \{ \mathbf{A}_{s}^{\mathcal{F}}, \mathbf{A}_{m}^{\mathcal{F}}, \mathbf{A}_{l}^{\mathcal{F}}\}$ and $\mathbb{A}^{\mathcal{G}} = \{ \mathbf{A}_{s}^{\mathcal{G}}, \mathbf{A}_{m}^{\mathcal{G}}, \mathbf{A}_{l}^{\mathcal{G}}\}$ are obtained. 
Likewise, sets of multi-scale feature maps $\mathbb{X}^{\mathcal{F}} = \{ \mathbf{X}_{s}^{\mathcal{F}}, \mathbf{X}_{m}^{\mathcal{F}}, \mathbf{X}_{l}^{\mathcal{F}}\}$ and $\mathbb{X}^{\mathcal{G}} = \{ \mathbf{X}_{s}^{\mathcal{G}}, \mathbf{X}_{m}^{\mathcal{G}}, \mathbf{X}_{l}^{\mathcal{G}}\}$ are also obtained. All elements of $\mathbb{X}^{\mathcal{F}}$ and $\mathbb{A}^{\mathcal{F}}$ are interpolated to the size of $\mathbf{I}_{m}$.
CAM of $k^{th}$ foreground class is further scaled as $\mathbf{A}^{k} \leftarrow \mathbf{A}^{k}/max(\mathbf{A}^{k})$ as in ~\cite{ahn2018learning,kweon2021unlocking}.

\subsection{Regional Contrastive Module (RCM)} \label{sec:RCM}
The proposed RCM is depicted in the right of Fig.~\ref{fig:method_framework}.
To provide pixel-wise guidance for the MainNet, RCM generates two forms of self-supervision.
First, from the CAMs of the SupportNet, the RCM generates class region masks that assign each pixel to the class.
Second, the RCM obtains the class-wise prototypes by averaging the features of the SupportNet within the class region mask of each class.
Such self-supervisions are provided to the MainNet in a contrastive manner.
Throughout the Sec.~\ref{sec:RCM}, for simplicity, we denote \textit{msinf}-CAMs of SupportNet and $\mathbf{X}_m$ as $\mathbf{A}$, $\mathbf{X}$, respectively.

\noindent \textbf{Class Region Masks}
Motivated by the observation that the CAMs have enough capability to localize the regions of each class even at the early stage of the training process~\cite{jiang2019integral,kweon2021unlocking}, we obtain a regional self-supervision using the CAMs from the SupportNet during the training.
We regard the CAM as a pixel-wise score map for being classified to that class.
For every pixel, by assigning value 1 according to the maximum argument of \textit{msinf}-CAMs from the SupportNet among foreground classes $C_{fg}$, we generate class region masks.
To revise the rough localization of CAMs to the reliable self-supervision for learning, we only consider pixels that have the activation higher than $threshold$ when generating the class region masks.

So, the class region mask of $k^{th}$ foreground class ($\mathbf{M}^k$) can be acquired from the $\mathbf{A}^\mathcal{G}$ as follows:
\begin{equation}
    \mathbf{M}^{k}_{p} = 
    \begin{cases}
        1,& \text{if} \hspace{5pt} k = \argmax_{{c}\in C_{fg}} \mathbf{A}^{\mathcal{G},c}_{p}\\ 
        &\hspace{12pt} and \hspace{5pt} \mathbf{A}^{\mathcal{G},k}_{p}>threshold\\
        0,& \text{otherwise}.
    \end{cases}
\end{equation}
where the $p$ denotes pixel position.
Note that the class region mask is obtained only for the existing classes in image-level GT.

Here, instead of making the networks directly predict the background with an additional channel, we regard the pixels with low foreground scores to be the background as: 
\begin{equation}
    \mathbf{M}^{0}_{p} = 
    \begin{cases}
        1,& \text{if} \hspace{5pt} {\mathbf{M}^{c}_{p}}=0 
        \hspace{5pt} \forall {c}\in C_{fg}\\ 
        0,& \text{otherwise}.
    \end{cases}
\end{equation}

Several studies exploit region masks for pseudo-labels for semantic segmentation~\cite{zhang2020reliability, wu2021embedded, van2021unsupervised}; however, the proposed method is advantageous in terms of the quality and the stability of the self-supervision.
Instead of using fixed sources of regional information such as the off-the-shelf saliency module~\cite{van2021unsupervised} or pre-trained classifier ~\cite{oh2021background}, we obtain region masks from the CAMs of SupportNet.
Moreover, we separate the network providing self-supervision for object localization (SupportNet) and the network learning from that guidance (MainNet), using EMA~\cite{He_2020_CVPR}.
This enables more stable training than the methods of which backbone is updated by self-supervision from itself~\cite{zhang2020reliability, wu2021embedded}.
Therefore, the acquired self-supervision is not only continually revised as training proceeds but also stably delivered to the MainNet.

\noindent \textbf{Class-wise Prototypes Generation (CPG)}
In the RCM, the class region masks not only serve as self-supervision for notifying which pixel should be classified to which class but also are utilized in Class-wise Prototype Generation (CPG) for obtaining the prototype of each class.
Even though pixel-level self-supervision is an effective way to impose spatial constraints with only image-level supervision, there still exists a chance for the self-supervision to be imprecise, compared to the pixel-level GT supervision annotated by the human.
Therefore, it is important to handle such unwanted derailment caused by imperfect self-supervision.
Instead of directly computing the cross-entropy loss between CAMs $\mathbf{A^{\mathcal{G}}}$ and the class region masks, we propose to apply feature-level contrastive learning as a soft criterion.
By averaging the SupportNet feature $\mathbf{X}^{\mathcal{G}}$ in accordance with class region mask of the $k^{th}$ class, we define the $k^{th}$ class prototype $\mathbf{pt}^{k} \in \mathbb{R}^{D}$ for each class as follows:
\begin{equation}
    \mathbf{pt}^{k} = 
    \sum_{p} \mathbf{M}^{k}_{p} \mathbf{X}^{\mathcal{G}}_{p}.
\end{equation}
We define the prototypes not only for foreground classes but also for a background class.
Here, the extracted prototypes represent the regions of existing classes on each image.
By averaging the prototypes from multiple images in a class-wise manner, we acquire mean version of the $k^{th}$ class-wise prototype $\overline{\mathbf{pt}}^{k} = \sum\mathbf{pt}^{k}$ and enables it to better represent the class.
\vspace{+3pt}

\noindent\textbf{Regional Contrastive Learning} 
With $\overline{\mathbf{pt}}$, pixel-wise features extracted from the MainNet ($\mathbf{X}^{\mathcal{F}}$) are trained in a contrastive manner.
We first map the features and prototypes on the $D$-dimensional hyper-sphere through L2 normalization.
Then, we define the similarity $sim(\cdot)$ between pixel-wise feature $\mathbf{X}_{p}^{\mathcal{F}}$ and prototype $\overline{\mathbf{pt}}^k$ with exponential function and dot product as follows:
\begin{equation}
    sim(\mathbf{X}_{p}^{\mathcal{F}},\overline{\mathbf{pt}}^k) = \exp{(\mathbf{X}_{p}^{\mathcal{F}} \cdot \overline{\mathbf{pt}}^k)}.
\end{equation}

The goal of the RCM is to make the pixel-wise feature of a certain class be close to the prototype of that class and far from the prototypes of the other classes.
Note that, unlike the prototypical contrastive learning~\cite{van2021unsupervised} that aims to make anchor features closer to the prototype of the same image than those of different images, we build class-wise prototype from multiple images and train the features to be close to the corresponding prototype.
Our loss function is based on mutual-information maximization as in~\cite{oord2018representation}.
In specific, the contrastive loss function of RCM for $k^{th}$ class with temperature $T$ is shown as follows:
\begin{equation}
    \mathcal{L}^{k}_{RCM} = -\sum_{p}{\mathbf{M}}^{k}_{p} \frac{sim(\mathbf{X}_{p}^{\mathcal{F}},\overline{\mathbf{pt}}^k/T)}
    {\sum^{N}_{w=0} sim(\mathbf{X}_{p}^{\mathcal{F}},\overline{\mathbf{pt}}^w/T)},
\end{equation}
and we sum the contrastive losses for all classes to compute $\mathcal{L}_{RCM}$.

\subsection{Multi-scale Attentive Module (MAM)}

Multi-scale inference is a widely used technique in semantic segmentation~\cite{zhao2017pyramid,takikawa2019gated,zhu2019improving,choi2020cars,sun2019high}, which merges the outputs inferred from multi-scale versions of the input image via pooling.
In the case of semantic segmentation, when a high resolution image is given as an input, fine-details are well captured while missing some global context.
On the contrary, when the input image has low resolution, the network can capture global context information well while lacking some fine-details~\cite{tao2020hierarchical}.
Therefore, by using the multi-scale inference technique, existing segmentation networks have been able to improve their performance during inference.
As the multi-scale inference approach can also relieve large variations in CAMs from multi-scale images, it has also been well-utilized in the field of WSSS~\cite{ahn2018learning,wang2020self,kweon2021unlocking,zhang2020reliability,lee2021anti,chang2020weakly} to generate pseudo-labels.
CAMs obtained from an image of a specific resolution contain meaningful information that is difficult to be obtained at the other resolutions, so the performance is greatly improved through the multi-scale inference technique.
In MAM, motivated by this, CAMs at each scale are guided to learn more from dissimilar CAMs of different scales while leveraging the high localization capability of \textit{msinf}-CAMs.
\begin{figure}[t]
    \centering
    \includegraphics[width=1.0\linewidth]{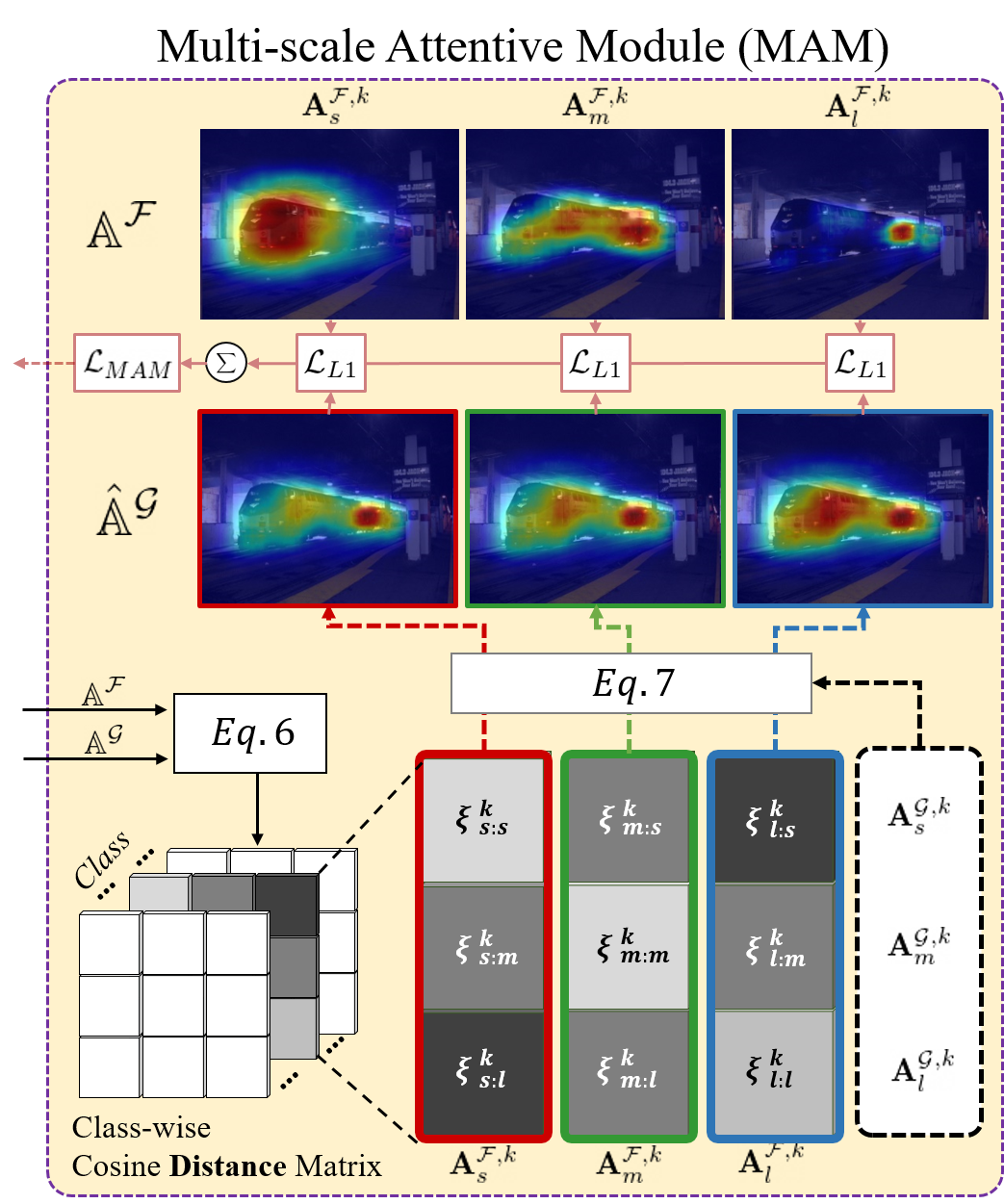}
    \caption{Overview of the proposed multi-scale attentive module (MAM). Loss is computed only for the existing classes in image-level GT. Here, $k$ is the \textit{train} class.}
    \label{fig:method_AML}
    \vspace{-5pt}
\end{figure}

The cosine distances between multi-scale CAMs from MainNet ($\mathbb{A}^{\mathcal{F}}$) and multi-scale CAMs from SupportNet ($\mathbb{A}^{\mathcal{G}}$) are calculated in a class-wise manner as shown in Fig.~\ref{fig:method_AML}. 
Here, cosine distance ($\xi$) is calculated only for classes that exist in image-level supervision as follows:
\begin{equation} \label{eq:dissim}
    \xi_{i:j}^k= 2-\frac{\mathbf{A}_{i}^{\mathcal{F},k}\cdot \mathbf{A}_{j}^{\mathcal{G},k}}{\left \| \mathbf{A}_{i}^{\mathcal{F},k} \right\|_{2}{\left \| \mathbf{A}_{j}^{\mathcal{G},k} \right \|}_{2}},
\end{equation} 
where $i,j\in\{s,m,l\}$ represent the scales and $\left \| \cdot  \right \|_{2}$  represents L2 norm.
Note that we vectorize $\mathbf{A}^{\mathcal{F}}$ and $\mathbf{A}^{\mathcal{G}}$ but it is omitted in Eq.~\ref{eq:dissim}.
The resulting dimension of the cosine distance matrix is $\mathbb{R}^{N\times 3 \times 3}$. The cosine distance matrix is used to create self-supervision for training multi-scale CAMs ($\mathbf{A}_{i}^{\mathcal{F}}$) obtained from MainNet.
The target CAMs ($\hat{\mathbf{A}}_{i}^{\mathcal{G}}$) are weighted summation of CAMs at various scales ($\mathbf{A}_{j}^{\mathcal{G}}$) according to the cosine distances.

The aforementioned process can be expressed as follows:
\begin{equation}
    \hat{\mathbf{A}}_{i}^{\mathcal{G},k} = \frac{1}{3} \sum_{j} \xi_{i:j}^{k} \cdot \mathbf{A}^{\mathcal{G},k}_{j}.
\end{equation}
The loss for MAM can be defined as:
\begin{equation}
    \mathcal{L}_{MAM}= \sum_{k} \sum_{i} \left \| \hat{\mathbf{A}}_{i}^{\mathcal{G},k}- \mathbf{A}_{i}^{\mathcal{F},k}    \right \|_{1},
\end{equation}
here, $\left \| \cdot  \right \|_{1}$ denotes L1 norm.
Note that only the MainNet $\mathcal{F}$ is updated by the loss function.
\vspace{-5pt}
\subsection{Loss Formulation}
\vspace{-5pt}
The proposed framework is trained with a combination of loss functions as follows:
\begin{equation}
    \mathcal{L}_{ours}= \lambda_{1}\mathcal{L}_{cls}+\lambda_{2}\mathcal{L}_{RCM}+\lambda_{3}\mathcal{L}_{MAM}.
\end{equation}
Here, $\mathcal{L}_{cls}$ denotes the binary cross entropy loss between the image-level class prediction from MainNet and the image-level classification label $\mathbf{t}$.
$\mathcal{L}_{RCM}$ and $\mathcal{L}_{MAM}$ represent the losses acquired from RCM and MAM, respectively.
$\lambda_1$, $\lambda_2$, and $\lambda_3$ are balancing parameters between the loss terms.

\begin{table}[t]
\begin{center}
\caption{Ablation study of the proposed framework.   
\textbf{SMM}: Single-Multi Module. \textbf{MMM}: Multi-Multi Module. \textbf{MAM}: the proposed Multi-scale Attentive Module. \textbf{RCM}: the proposed Regional Contrastive Module. The performance is evaluated on the PASCAL VOC \textit{train} set.}
\vspace{-10pt}
\label{tb:ablation_framework}
\resizebox{\linewidth}{!}{\renewcommand{\tabcolsep}{4pt}
\begin{tabular}{cccccccc}
\hline
Baseline  & SMM        & MMM         & MAM      & RCM & mIoU (\%)      & mIoU (w/ CRF, \%)\\\hline

\checkmark &           &           &           &     & 48.4    &54.4          \\\hline

\checkmark &\checkmark &           &           &     & 51.0    &56.7      \\\hline

\checkmark &           &\checkmark &           &     & 57.0    &62.2     \\\hline

\checkmark &           &           &\checkmark &     & 58.0    &63.4
\\\hline

\checkmark &           &           &           &\checkmark     & 57.6    &62.8      \\\hline

\checkmark &           &\checkmark &           &\checkmark     & 62.6 & 66.6 \\\hline
\checkmark &           &           &\checkmark &\checkmark     & \textbf{63.7} & \textbf{67.7} \\
\hline
\vspace{-25pt}
\end{tabular}}
\end{center}
\end{table}

\begin{figure}[t]
    \centering
    \includegraphics[width=1\linewidth]{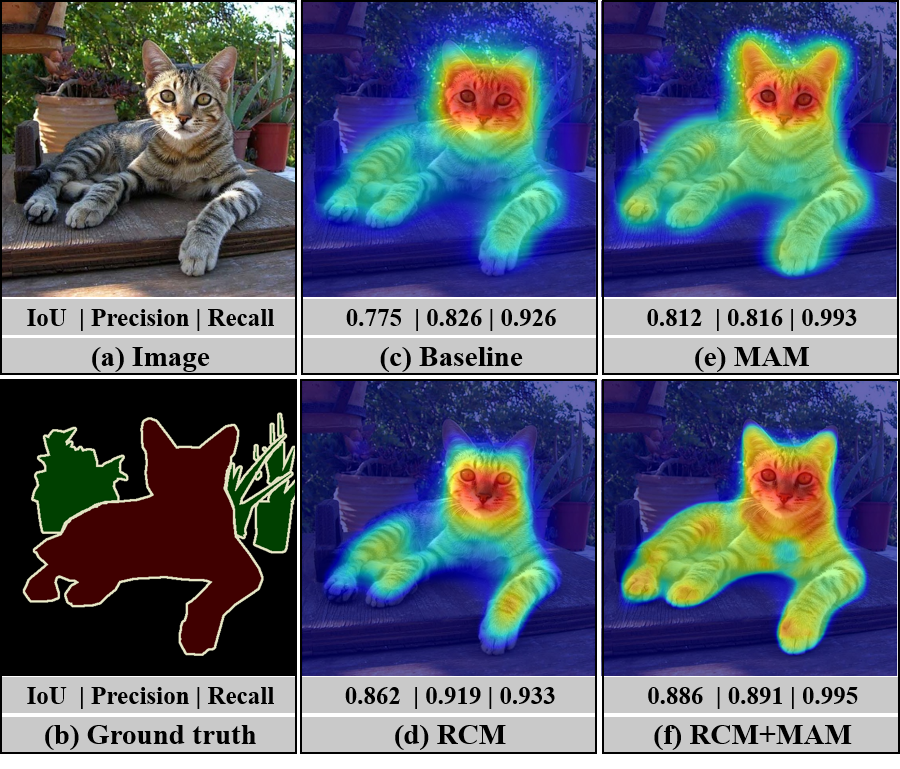}
    \vspace{-20pt}
    \caption{Qualitative comparison of CAMs with the proposed modules. IoU, precision, and recall for each CAM are also represented below. (a) and (b) represent input image and ground truth, respectively. From (c) to (f), baseline CAM, CAM with RCM, CAM with MAM, and CAM with both RCM and MAM.}
    \label{fig:experiment_ablation_cam}
    \vspace{-5pt}
\end{figure}

\section{Experimental Results}
\subsection{Dataset and Evaluation Metric}
We evaluate the proposed method on the PASCAL VOC 2012 dataset \cite{everingham2010pascal}. %and MS-COCO dataset~\cite{lin2014microsoft}.
PASCAL VOC 2012 dataset contains 21 categories including background.
As with other existing works, we train the proposed framework on the augmented training set (10,582) with image-level class labels only. 
For evaluation and comparison with other methods, we use validation (1,449) and test sets (1,456).  
We employ the mean Intersection over Union (mIoU) as an evaluation metric, which is a common standard in the field of WSSS.

\begin{table}[]
\begin{center}
\caption{Comparison of the mIoU, precision, and recall performances from the various settings in regards to the RCM and our proposed framework. \textbf{\textit{Direct}}: impose direct self-supervision to CAMs. \textbf{\textit{Cont}}: impose self-supervision in a contrastive manner. \textbf{W.S.}: weight-sharing. Differences between each experiment and baseline are noted below. \textbf{Bold} numbers represent the best results and \underline{underlined} numbers are the second-best ones. For each of the settings, the precision and recall performances are evaluated on the threshold that achieves the highest mIoU.}
\vspace{-5pt}
\label{tb:ablation_PrecisionRecall}
\renewcommand{\arraystretch}{1.4}
\resizebox{\linewidth}{!}{\renewcommand{\tabcolsep}{4pt}
\begin{tabular}{c|c|c|c|c|c}
\hline

Exp \# & \begin{tabular}[c]{@{}c@{}} RCM \\ 
self-sup. \end{tabular} & \begin{tabular}[c]{@{}c@{}} $\mathcal{G}$ Update \\
Policy \end{tabular} & \begin{tabular}[c]{@{}c@{}} mIoU (\%) \\ 
$\triangle$mIoU \end{tabular}   & \begin{tabular}[c]{@{}c@{}} Precision (\%)  \\ 
$\triangle$Precision \end{tabular}      & \begin{tabular}[c]{@{}c@{}} Recall (\%) \\ 
$\triangle$Recall \end{tabular}    \\ \hline

Baseline & -            & -             & 48.4        & 60.8         & 72.0    \\ \hline

1        & \textit{Direct}       & EMA           & \begin{tabular}[c]{@{}c@{}}52.3\\ 
(+3.9)\end{tabular}  & \begin{tabular}[c]{@{}c@{}}68.7\\ (+ 7.9)\end{tabular} & \begin{tabular}[c]{@{}c@{}}70.5\\ (-1.5)\end{tabular} \\ \hline

2        & \textit{Cont.} & W.S. & \begin{tabular}[c]{@{}c@{}}53.7\\ (+5.3)\end{tabular}  & \begin{tabular}[c]{@{}c@{}}68.9\\ (+8.1)\end{tabular} & \begin{tabular}[c]{@{}c@{}}72.1\\ (+0.1)\end{tabular} \\ \hline

\begin{tabular}[c]{@{}c@{}} 3 \\ (RCM) \end{tabular}  & \textit{Cont.}  & EMA            & \begin{tabular}[c]{@{}c@{}} 57.6\\ 
(+9.2)\end{tabular}  & \begin{tabular}[c]{@{}c@{}}\underline{72.3}\\ \underline{(+11.5)}\end{tabular} & \begin{tabular}[c]{@{}c@{}}74.0\\ (+2.0)\end{tabular} \\ \hline

\begin{tabular}[c]{@{}c@{}} 4 \\ (MAM) \end{tabular}  & -            & EMA            &                         \begin{tabular}[c]{@{}c@{}}\underline{58.0}\\ \underline{(+ 9.6)}\end{tabular}  & \begin{tabular}[c]{@{}c@{}}70.6\\ (+ 9.8)\end{tabular} & \begin{tabular}[c]{@{}c@{}}\underline{77.1}\\ \underline{(+5.1)}\end{tabular} \\ \hline

\begin{tabular}[c]{@{}c@{}} 5 \\ (RCM+MAM) \end{tabular} & \textit{Cont.}  & EMA            & \begin{tabular}[c]{@{}c@{}}\textbf{63.7}\\ \textbf{(+15.3)}\end{tabular} & \begin{tabular}[c]{@{}c@{}}\textbf{75.8}\\ \textbf{(+15.0)}\end{tabular} & \begin{tabular}[c]{@{}c@{}}\textbf{79.6}\\ \textbf{(+7.6)}\end{tabular} \\ \hline

\end{tabular}}
\end{center}
\vspace{-25pt}
\end{table}
\subsection{Implementation Details}
We employ ResNet38~\cite{wu2019wider} as a backbone network for both the MainNet and SupportNet with an additional intermediate 1$\times$1 convolution layer to extract feature maps ($\mathbf{X}$).
We set the channel dimension ($D$) of the feature maps for RCM as 256.
The input images are augmented by random cropping and resizing, horizontal flipping, color jittering~\cite{krizhevsky2017imagenet}.
The images are cropped in 192$\times$192 size and the resizing range is set between 0.5 and 1.3.
As in \cite{chen2017deeplab}, we use a poly learning rate policy that multiplies $(1-\frac{iter}{max\; iter})^{power}$ to the initial learning rate (0.01), where the power is set to 0.9.
We initially set the balancing parameters for loss as $\lambda_{1}=1$, $\lambda_{2}=1$, and $\lambda_{3}=0$.
After the impreciseness of the CAMs of the MainNet is sufficiently resolved with the RCM (epoch 6 in our implementation), we set $\lambda_{3}=1$ to address sparseness.
We set temperature ($T$) as 0.5, and the network is trained for 20 epochs with a batch size 16. 
The $threshold$ is set to 0.20 and detailed ablation regarding threshold will be discussed in the \textit{Appendix}. The momentum weight $\alpha$ is set to 0.997.
For the implementation of the mean prototype $\overline{\mathbf{pt}}^{k}$ of RCM, we average the prototypes along the mini-batch.
However, due to the limitation of batch size, in some mini-batches, several classes are missing and resulting in ``empty" prototypes.
To address this issue, we use the latest non-empty prototype for each class.
For the segmentation network, we use Deeplab~\cite{DBLP:journals/corr/ChenPKMY14} with ResNet38 backbone as in~\cite{ahn2018learning,kweon2021unlocking,sun2021ecs,zhang2021complementary,li2021pseudo}.

\subsection{Ablation studies} \label{ablation} 

\noindent \textbf{Component Analysis}
We design extensive ablation studies on the proposed modules (RCM and MAM) to discuss their roles in our framework.
To show the effectiveness of the proposed MAM, we conduct a comparison between three possible methods utilizing \textit{msinf}-CAMs: a single-multi module (SMM), a multi-multi module (MMM), and the proposed multi-scale attentive module (MAM). 
SMM and MMM are variations of the proposed MAM and similar in that they utilize the \textit{msinf}-CAMs from the SupportNet as self-supervision.
However, the MMM is not aided by the attention based on dissimilarity, unlike the proposed MAM. 
The SMM only uses single-scale CAMs $\mathbf{A}_{m}^{\mathcal{F}}$ when training the MainNet.
Table~\ref{tb:ablation_framework} shows the mean intersection over union (mIoU) of the CAMs with different settings.
All of the modules achieve higher performance compared to the baseline.
The performance of MMM is higher than that of SMM, and MAM achieves the highest performance (58.0\%) among the three with the help of attention.
The result of this ablation study shows that learning from CAMs with multiple and different scales is beneficial, and it would be even better if the module can notice how much the difference is.
It is also clear that the RCM greatly improves the performance of the CAMs from 48.4\% to 57.6\%.
The proposed framework with RCM and MAM achieves 63.7\%, which is a significant gain (15.3\%) compared to the baseline.

Qualitative comparison of CAMs in Fig.~\ref{fig:experiment_ablation_cam} visualizes roles of the modules and their complementarity.
RCM, a module designed to relieve the impreciseness by contrasting pixel-level features with class-wise prototypes, generates precise activation compared to the baseline.
MAM, on the other hand, covers less-discriminative regions according to the multi-scale attentive supervision based on \textit{msinf}-CAMs and dissimilarity, and thereby the resulting CAMs expand.
The proposed framework, along with the proposed two modules, generates CAMs that are sharp and fully highlight the entire object as well.
% Various examples are shown for other classes in Fig.~\ref{fig:quali_cam}.

\begin{table}[t]
\begin{center}
\caption{Performance (mIoU, \%) comparison with other state-of-the-art WSSS methods on the PASCAL VOC 2012 \textit{val} and \textit{test} set. Methods using only image-level supervision are listed in this table since we use neither saliency module nor external dataset at all. \textbf{Bold} numbers represent the best results.}
\vspace{-5pt}
\label{tb:voc_valtest}
\resizebox{\linewidth}{!}{\renewcommand{\tabcolsep}{8pt}
\begin{tabular}{l| c c  |c c}
\hline
\textit{Methods}         & \textit{Backbone}  &\textit{Pub.} & \textit{Val} & \textit{Test} \\ \hline \hline
AffinityNet \cite{ahn2018learning}       & ResNet38  &CVPR18 & 61.7  & 63.7   \\ 
ICD\cite {fan2020learning}               & ResNet101 &CVPR20 & 64.1  & 64.3        \\
IRNet \cite{ahn2019weakly}               & ResNet50  &CVPR19 & 63.5  & 64.8        \\ 
SSDD \cite{shimoda2019self}              & ResNet38  &ICCV19 & 64.9  & 65.5        \\ 
SEAM \cite{wang2020self}                 & ResNet38  &CVPR20 & 64.5  & 65.7        \\ 
Sub-category \cite{chang2020weakly}      & ResNet101 &CVPR20 & 66.1  & 65.9  \\
RRM \cite{zhang2020reliability}          & ResNet101 &AAAI20 & 66.3  & 66.5 \\ 
BES \cite{chen2020weakly}                & ResNet101 &ECCV20 & 65.7  & 66.6  \\
ECS~\cite{sun2021ecs}                    & ResNet38  &ICCV21 & 66.6 &67.6\\
AdvCAM~\cite{lee2021anti}                & ResNet101 &CVPR21 & 68.1 &68.0\\
OC-CSE\cite{kweon2021unlocking}          & ResNet38  &ICCV21 & 68.4 &68.2\\ CPN~\cite{zhang2021complementary}        & ResNet38  &ICCV21 & 67.8 &68.5\\
PMM~\cite{li2021pseudo}                  & ResNet38  &ICCV21 & 68.5 &69.0\\
\hline
Ours                                     & ResNet38  &-      &\textbf{69.7} &\textbf{70.9}             \\ \hline

\end{tabular}}
\end{center}
\vspace{-20pt}
\end{table}

\begin{table*}[t]
\begin{center}
\caption{Class-wise IoU comparison on PASCAL VOC 2012 \textit{val} set with  only image-level supervision.} %\vspace{-8pt}
\vspace{-10pt}
\label{tb:voc_val}
\resizebox{\textwidth}{!}{\renewcommand{\tabcolsep}{2pt}
\begin{tabular}{c|ccccccccccccccccccccc|c}
\hline
Method     &  {bkg} & {aero} & {bike} & {bird}& {boat} & {bottle}& {bus} & {car} & {cat}& {chair} & {cow} & {table} & {dog} & {horse} & {mbk} & {person} & {plant} & {sheep} & {sofa}& {train}& {tv}& {mIoU}\\ \hline\hline

AffinityNet \cite{ahn2018learning} &88.2 &68.2 &30.6 &81.1 &49.6 &61.0 & 77.8 &66.1 &75.1 &29.0 &66.0 &40.2 &80.4 &62.0 &70.4 &73.7 &42.5 &70.7 &42.6 &68.1 &\textbf{51.6} &61.7  \\ \hline

SEAM \cite{wang2020self} &88.8 &68.5 &33.3 &\textbf{85.7} &40.4 &67.3 & 78.9 &76.3 &81.9 &\textbf{29.1} &75.5 &48.1 &79.9 &73.8 &71.4 &\textbf{75.2} &48.9 &79.8 &40.9 &58.2 &53.0 &64.5  \\ \hline\

BES \cite{chen2020weakly} &88.9 &74.1 &29.8 &81.3 &53.3 &69.9 &\textbf{89.4} &\textbf{79.8} &84.2 & 27.9 &76.9 &46.6 &78.8 &75.9 &72.2 &70.4 &\textbf{50.8} &79.4 &39.9 &65.3 &44.8 &65.7 \\ \hline\hline

Ours &\textbf{91.0} &\textbf{83.8} &\textbf{35.0} &85.4 &\textbf{70.0} &\textbf{72.0} &87.4 &77.4 &\textbf{89.1} &27.7 &\textbf{81.9} &\textbf{48.7} &\textbf{84.3} &\textbf{82.0} &\textbf{74.7} &69.4 &49.3 &\textbf{83.4} &\textbf{47.2} &\textbf{74.7} &48.3 &\textbf{69.7}\\ \hline

\end{tabular}
}
\vspace{-15pt}
\end{center}
\end{table*}

\begin{figure*}[t]
    \centering
    \includegraphics[width=0.95\linewidth]{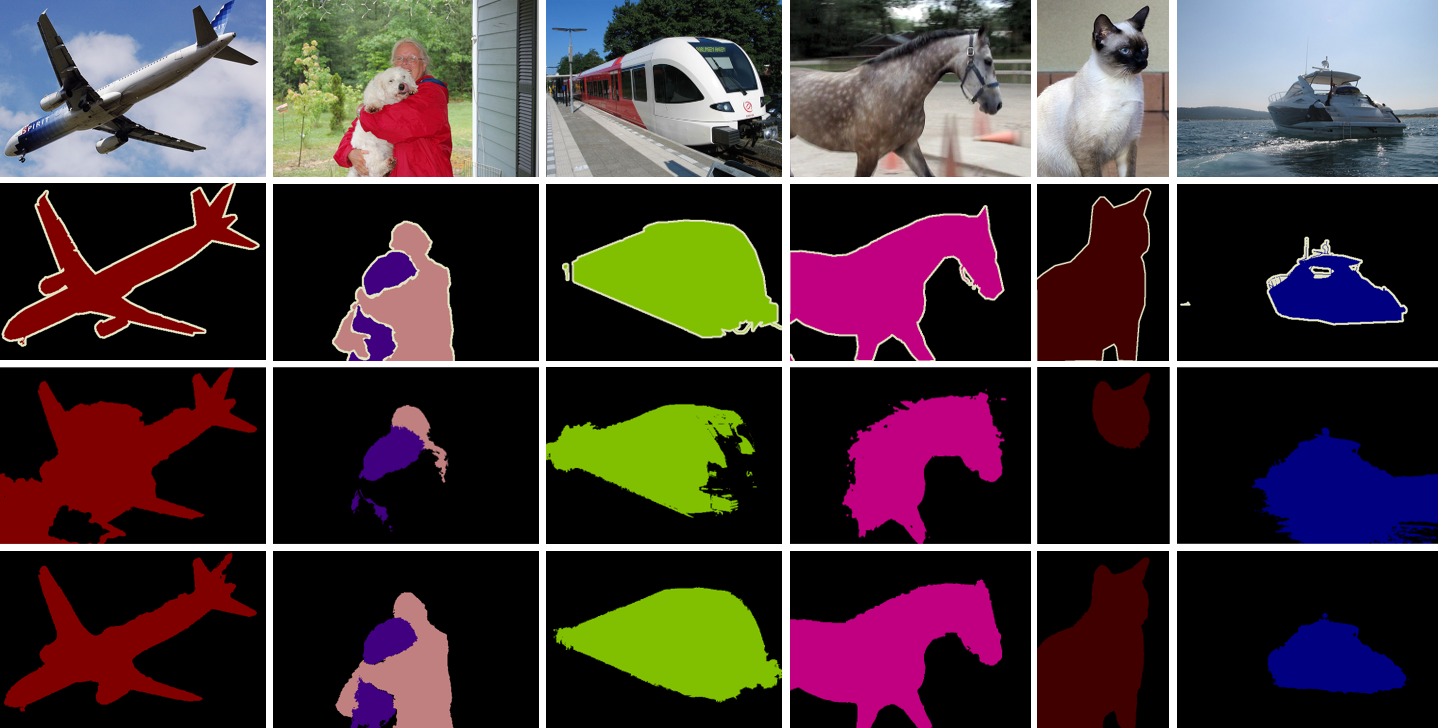}
    \caption{Qualitative comparison of the segmentation results on the PASCAL VOC 2012 \textit{validation} set. From top to bottom: Image, GT, Baseline~\cite{ahn2018learning}, and Ours. }
    \label{fig:experiment_dlcompare}
\vspace{-15pt}
\end{figure*}

To clarify the source of improvements, we also evaluate the precision and recall of the CAMs as in Table~\ref{tb:ablation_PrecisionRecall}.
Here, precision is the true activation over the whole activation while recall is the true activation over the GT map.
Precision and recall do not directly represent the preciseness and denseness of CAMs, but still can be effective measures to compare.
From the Exp \#1 to Exp \#3 in Table~\ref{tb:ablation_PrecisionRecall}, we conduct an ablation study on RCM. 
As shown in Exp \#1, when directly imposing self-supervision (in this case, class region mask only) to $\mathbf{A}^{\mathcal{F}}$, recall is decreased compared to the baseline.
Directly using self-supervision can increase mIoU and precision since it can provide pixel-level guidance, but reduces recall due to imperfection of self-supervision.
Comparing the result of Exp \#2 and Exp \#3, we can refer that EMA helps SupportNet provide more reliable supervision than weight sharing ($\mathbf{W.S.}$) when applying contrastive learning.
With the proposed RCM in Exp \#3, precision greatly increases compared to the baseline.
The performance gap between Exp \#1 and Exp \#3 shows the superiority of the proposed contrastive learning-based RCM, which is designed for WSSS.
In specific, unlike direct self-supervision, using contrastive learning to handle erroneous self-supervision increases the precision significantly, while not harming the recall.
Moreover, since MAM guides CAMs to be activated on even less-discriminative regions, Exp \#4 results in a increase in recall.
As shown in Exp \#5, applying both the RCM and MAM not only increases the mIoU by 15.3\%, but also provides the greatest benefit in terms of both precision and recall.
The result supports our design intention: RCM makes the model generate precise activation, and MAM helps the model highlight the whole object.

\subsection{Comparison with State-of-the-arts}

To further improve the quality of pseudo labels, we also apply a commonly used Random Walk (RW) approach~\cite{ahn2018learning} as in~\cite{wang2020self,chang2020weakly,lee2021anti,zhang2021complementary,sun2021ecs}. 
After applying the RW to our CAMs, the performance of pseudo labels achieve 71.0\% mIoU on PASCAL VOC 2012 \textit{train} set.
Then we train the Deeplab-LargeFOV \cite{DBLP:journals/corr/ChenPKMY14} with a ResNet38 backbone network with the corresponding pseudo labels. 

With the proposed framework, as shown in Table~\ref{tb:voc_valtest}, we achieve state-of-the-arts with 69.7\% and 70.9\% mIoU on PASCAL VOC 2012 \textit{val} and \textit{test} sets, respectively.
The proposed method outperforms the current state-of-the-art models by 1.9\% in the \textit{test} set, which is a significant margin.
Class-wise IoU comparison on PASCAL VOC 2012 \textit{val} set with only image-level supervision is also shown in Table~\ref{tb:voc_val}.
A qualitative comparison of the segmentation results between our method and baseline~\cite{ahn2018learning} is depicted in Fig.~\ref{fig:experiment_dlcompare}.

\subsection{Limitations}
Though the proposed framework achieves significant performance improvements compared to previous methods, our method also shares one common limitation with them.
As most WSSS methods utilize CAMs for generating pseudo-labels, the quality of pseudo-labels inevitably depends on the performance of the classifier.
When there is confusion between the classes (e.g., when both \textit{cow} and \textit{sheep} exist), the corresponding CAMs are also created confusingly. 
Since the proposed method generates class-wise prototypes based on CAMs, there exist a few cases where the resulting prototypes fail to fully represent the class causing confusion.
With a better performing classifier, we can expect the proposed method to obtain more representative prototypes as well as better self-supervision, and the performance would be more promising in the perspective of weakly supervised semantic segmentation.

\vspace{-5pt}
\section{Conclusion}
\vspace{-10pt}
In this paper, we proposed a novel framework deriving pixel-level self-supervision from image-level supervision to overcome the sparseness and impreciseness of CAMs.
Motivated by the promising representation capability of self-supervised learning and the necessity of pixel-level supervision, we devise region contrastive module (RCM) and multi-scale attentive module (MAM) that target impreciseness and sparseness, respectively.
With class region mask and class-wise prototype, by indirectly providing the pixel-level self-supervisions to the network in a contrastive manner, RCM resolves the problem of impreciseness in CAMs.
Multi-scale attentive module (MAM), on the other hand, aims to relieve the sparseness of CAMs by making CAMs expand to the less-discriminative regions based on \textit{msinf}-CAMs and dissimilarity between multi-scale CAMs.
By conducting extensive ablation studies, we also experimentally verify that the proposed RCM and MAM successfully resolve the problem of impreciseness and sparseness, respectively.
Using the semantic segmentation model trained with our pseudo-labels, we achieve state-of-the-art performance both on the
PASCAL VOC 2012 the \textit{validation} and \textit{test} set utilizing only the image-level classification labels in the WSSS with great margin.

%%%%%%%%% REFERENCES
{\small
\bibliographystyle{ieee_fullname}
\bibliography{egbib}
}

%%%%%%%%% APPENDIX
\newpage
\begin{appendix}

\section{Robustness to Scale-variance} 
\begin{figure}[h]
    \centering
    \includegraphics[width=1.0\linewidth]{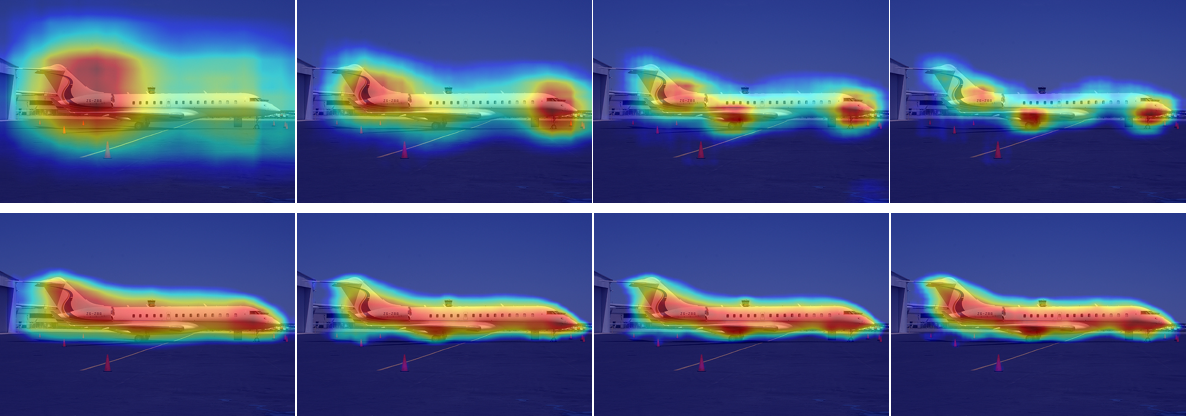}
    \vspace{-20pt}
    \caption{Qualitative comparison of CAMs with various input image scale. From left to right: image with ratio 0.5, 1.0, 1.5, and 2.0. From top to bottom: CAMs from the baseline, CAMs from our framework.}
    \label{fig:supple_quali_scale}
    \vspace{-5pt}
\end{figure}

Along with the two main problems of impreciseness and sparsity of CAMs, there exists one more limitation.
Since classification loss itself cannot impose spatial constraints, the resulting CAMs vary greatly depending on input image resolution (\textit{i.e.}, scale-variance)~\cite{wang2020self}. On the contrary, the proposed framework alleviates the scale-variance of CAMs with the pixel-level self-supervision from RCM and MAM. 
As shown in Table~\ref{tb:supple_multi_scale}, the proposed framework produces much less scale-variant CAMs than the baseline by a large margin.
Moreover, compared to the CAMs of SEAM~\cite{wang2020self} which also aims to minimize scale-variance, our framework produces more scale-consistent CAMs.
We quantitatively show that leveraging the pixel-level self-supervision from RCM and MAM helps CAMs to be much more scale invariant.
Qualitative comparison of CAMs is also shown in Fig.~\ref{fig:supple_quali_scale} with various image ratio (0.5,1.0,1.5,2.0), showing that CAMs from the proposed framework much less scale-variant than those of the baseline.

\begin{table}[h]
\begin{center}
\caption{Performance comparison on various single-scale CAMs and multi-scale inferred CAMs (\textit{msinf}-CAMs). $\mu$ and $\sigma$ stand for mean and standard deviation of single-scale CAMs, respectively.} 
\vspace{-5pt}
\label{tb:supple_multi_scale}
\resizebox{\linewidth}{!}{\renewcommand{\tabcolsep}{6pt}
\begin{tabular}{c|c|c|c}
\hline
Inference scale            & Baseline (mIoU)& SEAM~\cite{wang2020self} (mIoU)  & Ours (mIoU) \\ \hline
{[0.5]}               & 40.2\%         & 49.4\%  & 60.8\%  \\
{[1.0]}               & 47.0\%         & 51.6\% & 62.8\%  \\
{[1.5]}               & 49.0\%         & 52.3\% & 61.7\%  \\
{[2.0]}               & 47.9\%         & 49.8\% & 59.7\%  \\ \hline
$\mu \pm \sigma$      & 46.0$\pm$3.4\% & 50.7$\pm$1.2\%   & 61.2$\pm$1.1\%  \\
{[0.5, 1.0, 1.5, 2.0]}& 48.4\%         & 55.41\%    & 63.7\% \\ \hline
\end{tabular}}
\vspace{-10pt}
\end{center}
\end{table}

\begin{figure}[t]
    \centering
    \includegraphics[width=1.0\linewidth]{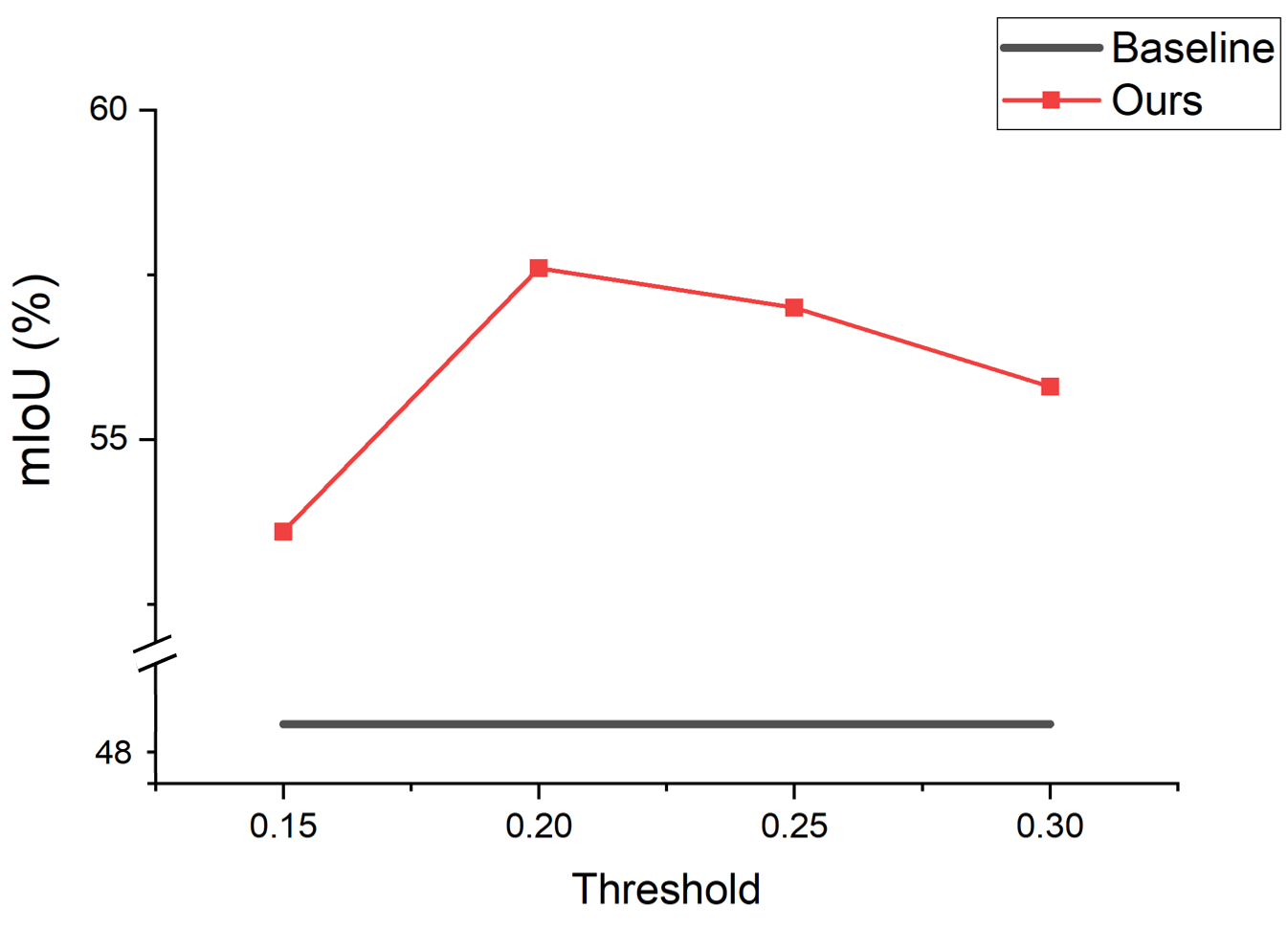}
    \caption{A plot of mIoU performances over \textit{threshold} in our framework with RCM. The red line represents the proposed method, while the black line represents the baseline.}
    \label{fig:supple_threshold}
    \vspace{-10pt}
\end{figure}

\section{Experiment Regarding \textit{threshold}}
As mentioned in the main paper, we revise the rough localization of CAMs to acquire the reliable self-supervision for learning.
Here, we only consider pixels that have their activation higher than the $threshold$ when generating the class region masks as in the Eq.~\textcolor{red}{1} of the main paper.
We set the $threshold$ as 0.20 for the results in the main paper. 
For the selection of $threshold$, we observe CAMs from the model trained with classification loss only and choose a qualitatively plausible threshold (0.20$\sim$0.25) for the CAMs.

To verify the sensitivity of our framework over the different \textit{threshold}s, we conduct an experiment while adjusting it.
Figure~\ref{fig:supple_threshold} shows the resulting performance of the CAMs from the proposed framework trained with the different \textit{threshold}s.
It shows that the proposed framework outperforms the baseline regardless of the \textit{threshold}.
Compared to the stable performance change at higher threshold, it shows a relatively sensitive performance change at the lower threshold.
In our perspective, this tendency is caused by the involvement of less confident foreground regions of CAMs, which can degrade the quality and stability of self-supervision. 
To relieve such erroneous self-supervision, we use the contrastive learning in the RCM and help our framework be constantly superior over the baseline.

\section{Network Architecture}
\begin{figure}[h]
    \centering
    \includegraphics[width=1.0\linewidth]{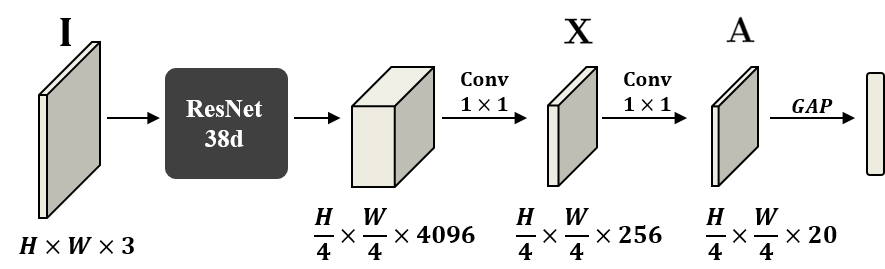}
    \caption{Network architecture detail. Both MainNet and SupportNet have identical architecture. Below each feature map, the dimension of the corresponding feature is written. Notation follows the main paper. Here, $H, W$ denotes height and width of the given image, respectively. } 
    \label{fig:supple_architecture}
    \vspace{-15pt}
\end{figure}

As with many other previous weakly supervised semantic segmentation (WSSS) approaches~\cite{ahn2018learning,shimoda2019self,wang2020self,chang2020weakly,zhang2020reliability,kweon2021unlocking}, we employ the ResNet38~\cite{wu2019wider} as the backbone for both the MainNet and the SupportNet.
To extract feature maps ($\mathbf{X}$) for class-wise prototypes and pixel-wise contrastive learning, we add an intermediate 1$\times$1 convolution layer. With the additional 1$\times$1 convolution layer, the CAMs ($\mathbf{A}$) can be acquired.
We set the channel dimension ($D$) of the feature maps to be 256.
The architecture of the network is shown in Fig.~\ref{fig:supple_architecture} with the intermediate feature maps and their corresponding dimensions.

\section{Additional Qualitative Comparison}

\begin{figure*}
    \centering
    \includegraphics[width=0.94\linewidth]{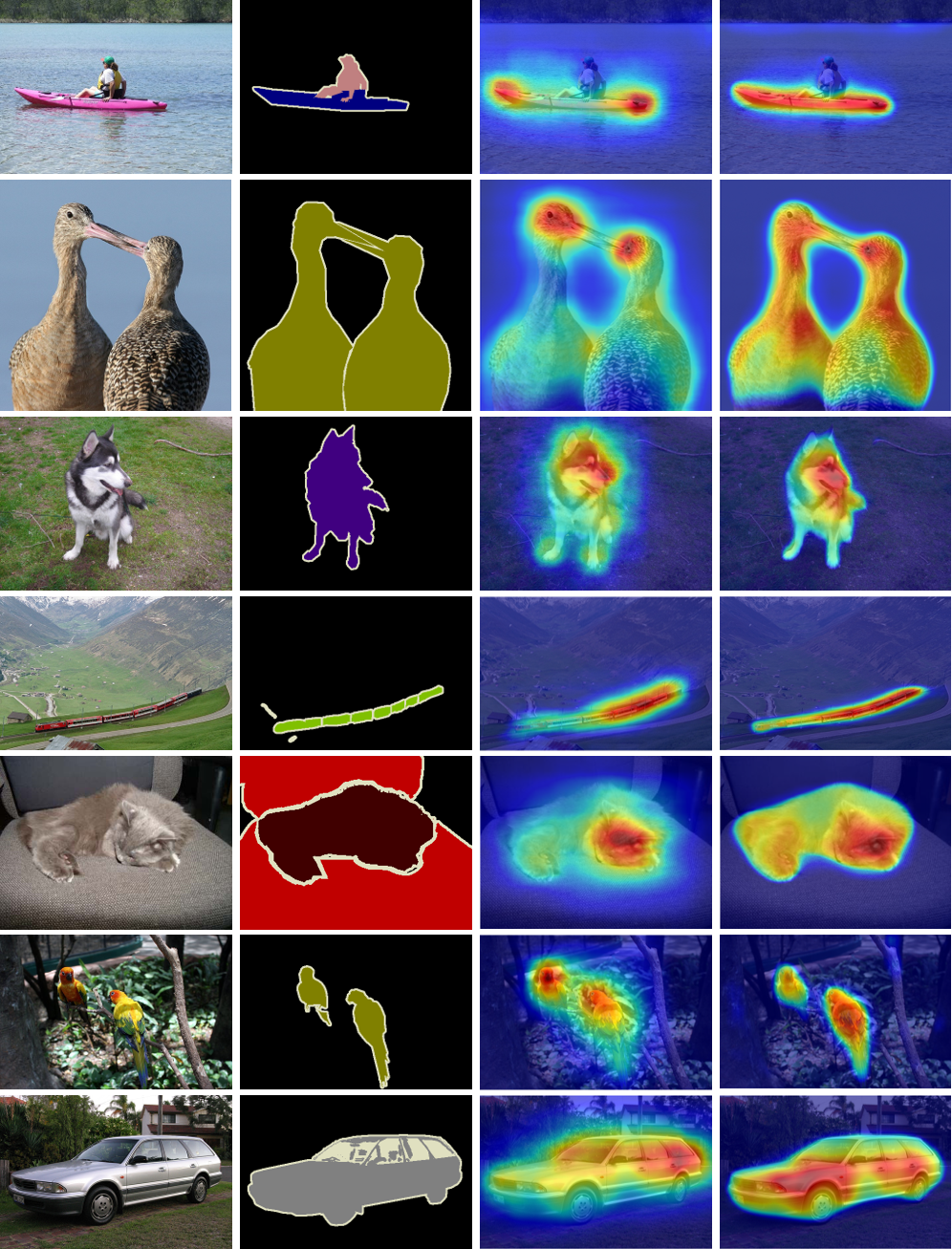}
    \caption{Qualitative comparison of CAMs on PASCAL VOC 2012 \textit{train} set. From left to right: images, ground-truths, baseline CAMs with ResNet38 backbone, our CAMs. }
    \label{fig:supple_qualicam}
    \vspace{-5pt}
\end{figure*}

\begin{figure*}
    \centering
    \includegraphics[width=0.96\linewidth]{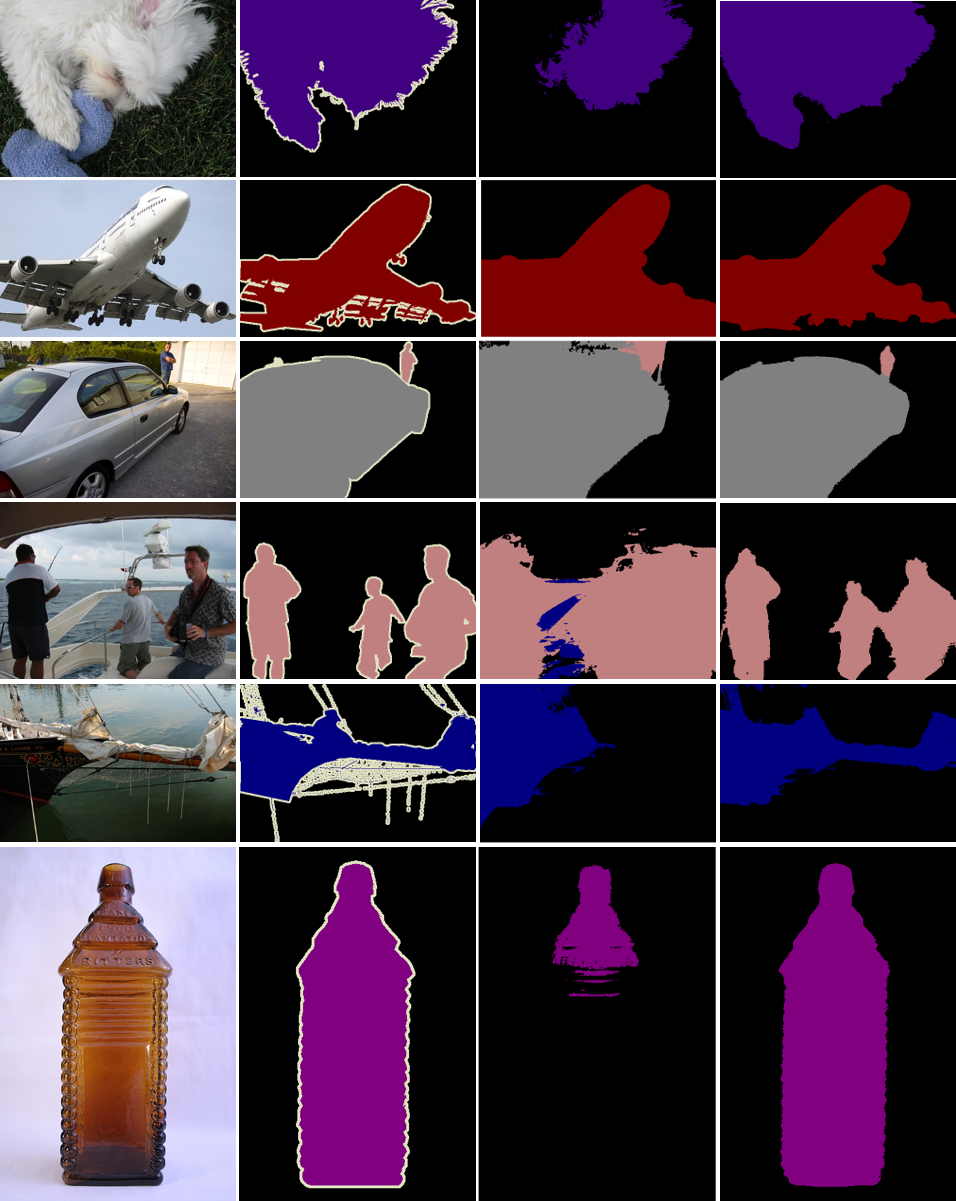}
    \caption{Qualitative comparison of semantic segmentation maps on PASCAL VOC 2012 \textit{validation} set. From left to right: images, ground-truths, Deeplab with the baseline~\cite{ahn2018learning}, Deeplab by ours. }
    \label{fig:supple_qualidl}
    \vspace{-5pt}
\end{figure*}

To show that our proposed framework not only relieves the impreciseness of CAMs but also addresses the sparseness by expanding CAMs to less-discriminative regions, qualitative comparison of CAMs with baseline that are not included in the main paper due to page limit are shown in Fig.~\ref{fig:supple_qualicam}. 
With the generated pixel-level pseudo-labels by applying RW~\cite{ahn2018learning} to CAMs, we train the semantic segmentation network~\cite{DBLP:journals/corr/ChenPKMY14} and achieve \textit{state-of-the-art} on WSSS using only image-level supervision. Additional qualitative comparison of segmentation maps is shown in Fig.~\ref{fig:supple_qualidl}. 

\section{Reproducibility}
For the reproducibility of our work, we also share our PyTorch implementations.
Due to the memory limitation of Microsoft CMT (maximum 100MB), we could not provide pre-trained checkpoint of our framework (about 400MB).
Instead, we upload our \textit{train}, \textit{inference} and \textit{evaluation} code with README file for reviewers who want to reproduce our results. 

\end{appendix}

\end{document}